\documentclass[runningheads,a4paper]{llncs}

\usepackage{amssymb}
\usepackage{graphicx}

\usepackage{url}
\urldef{\mailsa}\path|{kevinleezhang, mohitsharma, jackyliang, okroemer}@cmu.edu|    
\newcommand{\keywords}[1]{\par\addvspace\baselineskip
\noindent\keywordname\enspace\ignorespaces#1}

\begin{document}

\mainmatter  

\title{A Modular Robotic Arm Control Stack for Research: Franka-Interface and FrankaPy}

\titlerunning{Franka-Interface and FrankaPy}

%
%

\author{
Kevin Zhang\footnote[1]{
Equal Contribution.
This work was supported by Sony AI, the NSF Graduate Research Fellowship Program Grant No. DGE 1745016, and the Office of Naval Research Grant No. N00014-18-1-2775.
}
\and Mohit Sharma$^\star$ 
\and Jacky Liang$^\star$ 
\and Oliver Kroemer}
\authorrunning{Franka-Interface and FrankaPy}

\institute{
Intelligent Autonomous Manipulation Lab \\
Robotics Institute, Carnegie Mellon University,\\
5000 Forbes Ave. Pittsburgh, PA, USA\\
\mailsa\\
\url{https://labs.ri.cmu.edu/iam/}
}

%
%

\toctitle{Franka-Interface and FrankaPy}
\tocauthor{Zhang, Sharma, Liang, and Kroemer}
\maketitle

\begin{abstract}

We designed a modular robotic control stack that provides a customizable and accessible interface to the Franka Emika Panda Research robot.
This framework abstracts high-level robot control commands as skills, which are decomposed into combinations of trajectory generators, feedback controllers, and termination handlers.
Low-level control is implemented in C++ and runs at $1$kHz, and high-level commands are exposed in Python.
In addition, external sensor feedback, like estimated object poses, can be streamed to the low-level controllers in real time.
This modular approach allows us to quickly prototype new control methods, which is essential for research applications.
We have applied this framework across a variety of real-world robot tasks in more than $5$ published research papers.
The framework is currently shared internally with other robotics labs at Carnegie Mellon University, and we plan for a public release in the near future.

\keywords{Robot Control, Manipulation, Control System}
\end{abstract}
\begin{figure}[ht]
    \centering
    \includegraphics[width=\textwidth]{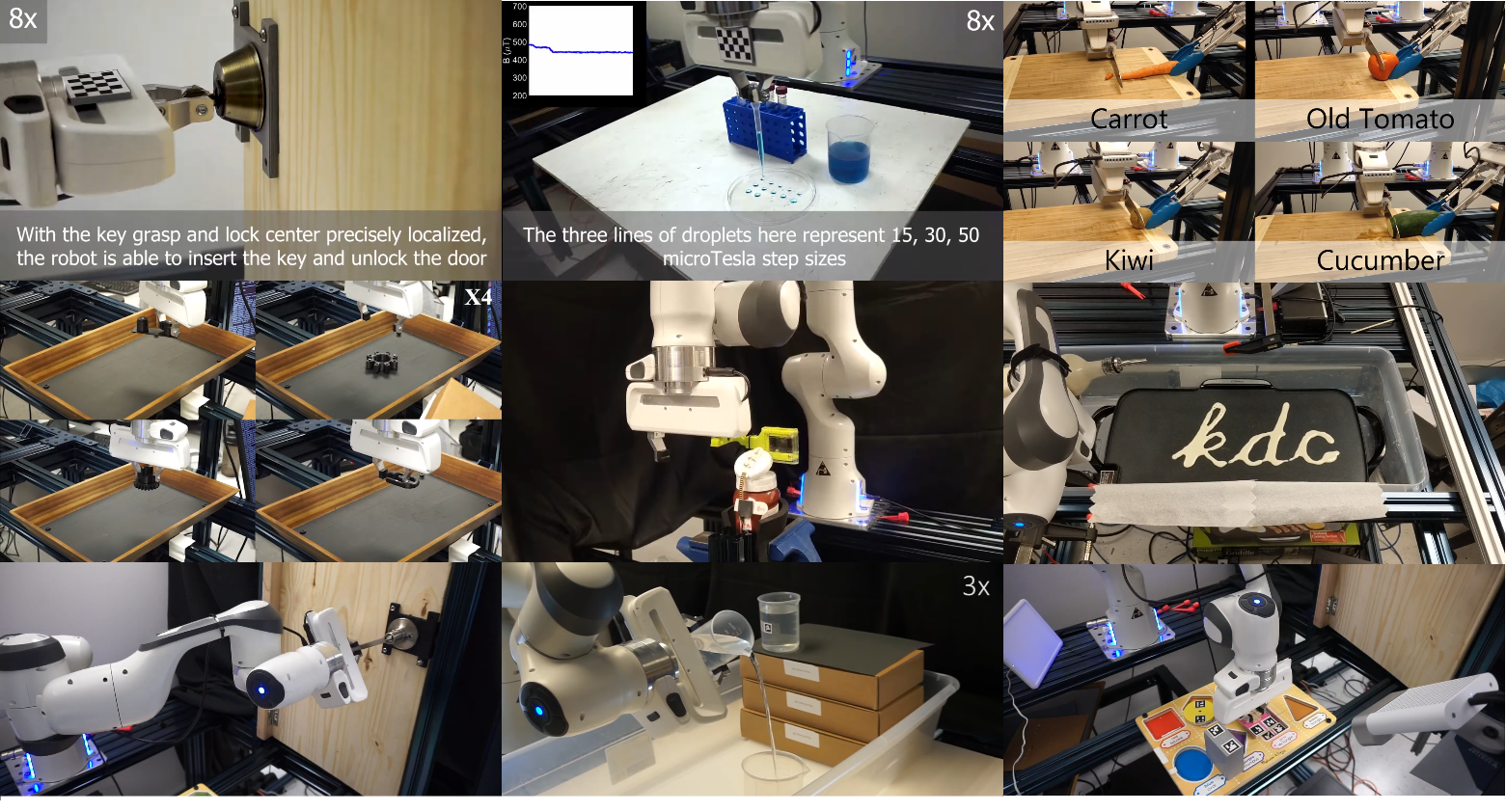}
    \caption{Collage of Robot Tasks implemented using Franka-Interface and FrankaPy. From left-to-right top-to-bottom, they include unlocking a door with a key, pipetting, cutting vegetables, picking up 3d printed versions of industrial parts, opening a ketchup bottle cap, writing letters using pancake batter, opening a door, pouring water from a beaker, and peg-in-hole insertion.}
    \label{fig:franka-interface-collage}
\end{figure}

\section{Introduction}

To facilitate robot programming, commercial collaborative robot arms (cobots) often ship with simple drag-and-drop programming interfaces or teach pendants.
These interfaces allow an operator to perform teleoperation, kinesthetic teaching, or hardcoding waypoints to program robot behavior, and they have allowed robots to perform tasks in areas such as manufacturing with minimal setup time~\cite{villani2018survey}.
However, such robot interfaces tend to be very GUI and operator dependant, which do not align well with the requirements of research development, where the ability to rapidly prototype with programming languages such as Python and C++ and incorporating outside sensor feedback is highly desirable. 
Companies like Universal Robots have catered to their academia audience by developing a simple way for researchers to interface with their robots using Robot Operating System (ROS)~\cite{quigley2009ros} and MoveIt \cite{chitta2012moveit}, through which a programmer is able to send joint position and velocity commands.
However, their ROS package~\cite{universal_robot} is unable to support access to the robot's full capabilities such as the UR5e's force control mode.

In this technical report, we describe Franka-Interface and Frankapy, a modular robot control software using C++ and Python that allows researchers to quickly prototype robot controllers on the Franka Emika Panda Research Robot\footnote{\url{https://www.franka.de/}}.
Franka-Interface uses libfranka~\cite{libfranka} to interface with the Franka robot, and Frankapy provides Python APIs for Franka-Interface.
Our software stack allows users to quickly implement control schemes such as Cartesian Impedance and LQR controllers in C++, all while maintaining an easy-to-use Python interface.
In addition, our software allows easy access to the robot's states and provides sensor feedback to controllers during skill execution.
Finally, we ensure the ability to control the Franka at $1$kHz and quickly change between different controllers.

\section{Comparison to franka\_ros and ros\_control}

Franka Emika has developed the franka\_ros \cite{franka_ros} package using ros\_control \cite{chitta2017ros_control} that exposes the majority of their libfranka API to control their Panda Research robot at 1kHz. 
There are two primary limitations with franka\_ros.
One, the package is heavily dependant on ROS and ros\_control, which means robot controllers implemented here have to conform to ros\_control guidelines. 
This makes it difficult to add additional functionality on the robot controllers, such as accepting external sensor inputs in real-time to guide the robot's trajectory.
Second, ros\_control's MultiInterfaceController class only allows $4$ robot hardware controllers to be loaded at a time, while libfranka has a total of $9$ robot controllers that can be used to command the robot. 
For example, if the software is initially loaded the Joint Position, Joint Velocity, Joint Torque, and Cartesian Pose robot controllers in one MultiInterfaceController, but the user also needs to use the Cartesian Velocity robot controller for another part of the task, the previous controller would need to be stopped completely before the new controller can be launched.
This creates an undesirable time gap for switching robot controllers.

By contrast, our Franka-Interface allows us to switch between any of the 9 robot controllers present in libfranka.
Our controllers are able to process real-time sensor feedback from ROS and the Python client.
Our software stack also allows asynchronous Joint Position and Cartesian Pose inputs from a sensor publisher that allows MoveIt integration.
Finally, we are able to log robot states at 1kHz during skill executions.
We will explain our entire system pipeline in the following sections.
\section{System Overview}

\begin{figure}[ht]
    \centering
    \includegraphics[width=\textwidth]{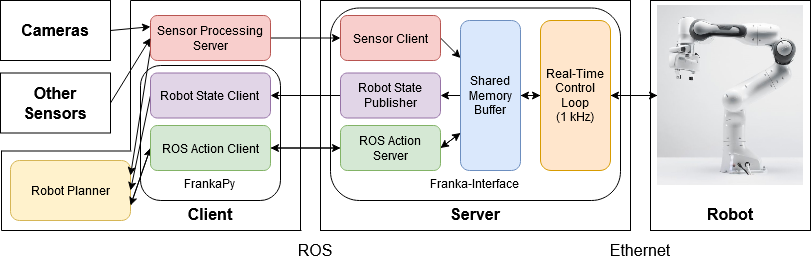}
    \caption{Franka Robot System Diagram}
    \label{fig:franka_system_diagram}
\end{figure}

Our system, shown in Figure \ref{fig:franka_system_diagram}, consists of three main parts: the Franka robot, a Server computer that is running a real-time Ubuntu Kernel, and a Client computer with a powerful CPU and GPU that handles the sensor processing as well as the high-level planning. 
While Server and Client can be the same computer, they are separate for use cases where a deep neural network is involved.
This is because libfranka, which runs on the server, requires a real-time kernel, and Nvidia drivers, required for running neural networks, are not compatible with real-time kernels.
If the Server and Client are separate computers, you can connect them either through ethernet or over wifi.
We will walk through the entire system starting from the Client on the left and slowly moving to the robot on the right.

\subsection{Client}

On the Client, we have the high-level robot planner, sensor processing servers, and FrankaPy which includes a robot state client and a ROS action client.
In addition, the Client is connected to the various cameras and other sensors in the robot setup. 
First, the high-level planner is typically a Python program that acts as a finite state machine, and it takes processed sensor information as well as the current robot state as inputs to command the robot. 
Next, the sensor processing server typically consists of a neural network that is performing object recognition or pose tracking, the results of which are published continuously.
Both the sensor client on the Server and the high-level planner can subscribe to the results. 

The FrankaPy robot state client on the Client reads information about the robot's current joint positions, end-effector pose, experienced torques, current skill executing, etc. 
Finally, the FrankaPy ROS Action Client is a Python API that contains all of the skills that we can have the robot execute. 
Some example functions that the Frankapy wrapper contains are 'go\_to\_pose', 'go\_to\_joints', 'open\_gripper', and 'execute\_dmp'. 
These functions wrap around the standard ROS Action server goals, which contain the parameters for each skill that are serialized using Protocol Buffers \cite{protobuf}. 
They are sent over the ROS networking protocol to the ROS Action server on the Real-Time Server that receives and parses the parameters.

\subsection{Server}

\begin{figure}[ht]
    \centering
    \includegraphics[width=\textwidth]{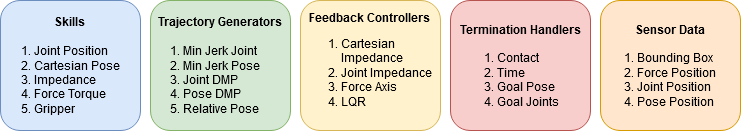}
    \caption{Examples of skill, trajectory generator, feedback controller, termination handler, and sensor types that are implemented in Franka-Interface.}
    \label{fig:franka-interface-types}
\end{figure}

The ROS action server on the Real-Time Server shares with the $1$kHz real-time control loop a common memory buffer.
This allows us to efficiently pass data, like controller parameters, between these two processes without costly serialization and deserialization.
The Real-Time Control Loop continually queries the shared memory buffer to see if a new skill's parameters are present.
When it sees that a new skill is available, it starts executing the skill. 
The skill parameters contain 5 main fields: the type of the skill, trajectory generator, feedback controller, termination handler, and the sensor topics to subscribe to. 
Franka-Interface allows us to mix and match the trajectory generator, feedback controller, termination handler, and sensor data manager for each skill depending on the task.
Figure \ref{fig:franka-interface-types} illustrates some commonly used types for each field, but this is not an extensive list of what our framework offers.

Examples of skill types are: using the robot's internal joint controller to go to joint positions, using the robot's internal pose controller to go to a robot pose, using impedance control to go to a robot pose, exerting forces in certain directions, moving the gripper, etc.
Next, we implemented several different trajectory generators for each skill type. 
For example, we implemented minimum jerk joint and robot pose trajectory generators that reduce the wear and tear on the robot. 
In addition, we implemented a special Dynamic Movement Primitives (DMP) trajectory generator that we used to have the robot imitate human kinesthetic demonstrations. 

For feedback controllers, we implemented a Cartesian impedance feedback controller that commands robot end-effector poses via a spring-damper system.
Otherwise, most of the other feedback controllers simply set the internal robot controller's Cartesian and joint impedances. 
However, others using this framework have been able to implement an LQR controller as well.
Finally, we have the termination handlers and the sensor topic parameters. 
There are time-based termination handlers that stop the skill after the designated amount of time has elapsed, goal-based termination handlers that tell a skill to stop when the robot is within a threshold of the final desired goal pose or joint configuration, and contact-based ones that stop based on force feedback.
The sensor topic parameters tell the sensor ROS client which topics to subscribe to and store message data in the shared memory for the real-time control loop to query. 

The last and most important part of the system is the real-time control loop, which communicates directly with the robot at a $1$kHz frequency. 
This loop handles commanding the robot over the libfranka C++ API as well as storing the robot state into logs and shared memory that can be accessed and published by the Robot State Publisher. 
We ensure the reliability of the real-time control loop by using multiple mutexes and shared memory partitions.

\subsection{Additional Functionality}

Some additional features we have implemented are: the ability to control multiple robots at once, reconfigurable virtual walls that act as safety barriers, and simple box collision checking. 
Franka\_ros also can launch multiple robots at once; however, they only currently support joint torque controllers.
We utilize franka\_ros's robot configuration file in order to visualize the robot's movements in real-time in Rviz \cite{quigley2009ros}. 

\subsection{Dependencies}

Our Franka-Interface and FrankaPy does depend on several packages such as Autolab\_Core \cite{autolab_core} for rigid transformations, Google Protocol Buffers \cite{protobuf} to serialize skill parameters and robot state data, and ROS \cite{quigley2009ros} for communication between the Client and the Server. 
In the future, for ease-of-use, we are planning to phase out ROS dependencies and switch to ZeroMQ \cite{zeromq} to handle communications between the Client and the Server.
\section{Impact}

\begin{figure}[ht]
    \centering
    \includegraphics[width=\textwidth]{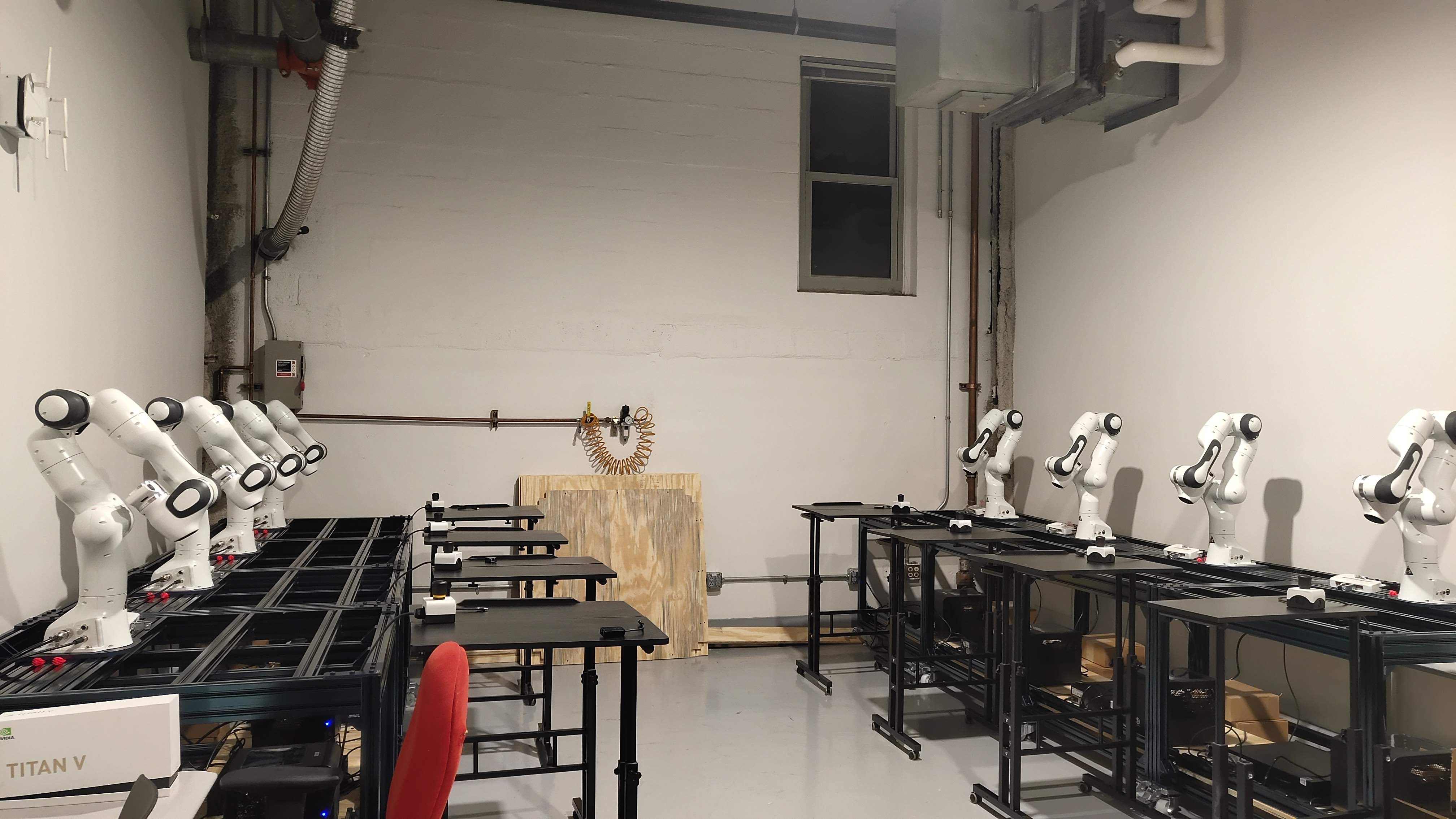}
    \caption{8 Franka Robot Setups used for a CMU RI course in Spring 2020.}
    \label{fig:franka_robot_setups}
\end{figure}

We have been using Franka-Interface and FrankaPy internally in the Intelligent Autonomous Manipulation lab since October 2018.
So far, 5 published papers from our lab have used this software for robot experiments as shown in Figure \ref{fig:franka-interface-collage} \cite{sharma2019learning,zhao2019towards,zhang2019leveraging,liang2020learning,hellebrekerslocalization,lagrassa2020learning}.
Recently, we have shared the software with other labs at the CMU Robotics Institute.
In addition, we have been able to use our package successfully for the Robot Autonomy class at the RI.
In the class, $50$ students were using the $8$ shared robots shown in Figure \ref{fig:franka_robot_setups} for lab experiments, like motion planning with obstacle avoidance.
We plan to publicly release the software once we resolve pending licensing issues and sufficiently improve code documentation.

\bibliographystyle{IEEEtran.bst}
\bibliography{bib}

\end{document}